\documentclass{article}

\usepackage{microtype}
\usepackage{graphicx}
\usepackage{subfigure}
\usepackage{booktabs} % for professional tables
\usepackage{enumitem}
\usepackage{amsmath}
\usepackage[none]{hyphenat}
\usepackage{flushend}

\usepackage{hyperref}

\usepackage[accepted]{icml2021}

\icmltitlerunning{Unified Shapley Framework to Explain Prediction Drift}

\begin{document}

\twocolumn[
\icmltitle{Unified Shapley Framework to Explain Prediction Drift}

% It is OKAY to include author information, even for blind
% submissions: the style file will automatically remove it for you
% unless you've provided the [accepted] option to the icml2021
% package.

% List of affiliations: The first argument should be a (short)
% identifier you will use later to specify author affiliations
% Academic affiliations should list Department, University, City, Region, Country
% Industry affiliations should list Company, City, Region, Country

% You can specify symbols, otherwise they are numbered in order.
% Ideally, you should not use this facility. Affiliations will be numbered
% in order of appearance and this is the preferred way.
\icmlsetsymbol{equal}{*}

\begin{icmlauthorlist}
\icmlauthor{Aalok Shanbhag}{equal,to}
\icmlauthor{Avijit Ghosh}{equal,to,goo}
\icmlauthor{Josh Rubin}{equal,to}
\end{icmlauthorlist}

\icmlaffiliation{to}{Fiddler Labs, Palo Alto, USA}
\icmlaffiliation{goo}{Northeastern University, USA}

\icmlcorrespondingauthor{Aalok Shanbhag}{aalokshanbhag@gmail.com}

% You may provide any keywords that you
% find helpful for describing your paper; these are used to populate
% the "keywords" metadata in the PDF but will not be shown in the document
\icmlkeywords{Machine Learning, ICML}

\vskip 0.3in
]

% this must go after the closing bracket ] following \twocolumn[ ...

% This command actually creates the footnote in the first column
% listing the affiliations and the copyright notice.
% The command takes one argument, which is text to display at the start of the footnote.
% The \icmlEqualContribution command is standard text for equal contribution.
% Remove it (just {}) if you do not need this facility.

% \printAffiliationsAndNotice{}  % leave blank if no need to mention equal contribution

\printAffiliationsAndNotice{\icmlEqualContribution{}} % otherwise use the standard text.

\begin{abstract}
Predictions are the currency of a machine learning model, and to understand the model's behavior over segments of a dataset, or over time, is an important problem in machine learning research and practice. There currently is no systematic framework to understand this drift in prediction distributions over time or between two semantically meaningful slices of data, in terms of the input features and points. We propose GroupShapley and GroupIG (Integrated Gradients), as axiomatically justified methods to tackle this problem. In doing so, we re-frame all current feature/data importance measures based on the Shapley value as essentially problems of distributional comparisons, and unify them under a common umbrella. We axiomatize certain desirable properties of distributional difference, and study the implications of choosing them empirically.

\end{abstract}

\section{Introduction}
Fundamental Machine Learning theory is structured around the assumption that the training and test data belong to the same distribution. While this is a reasonable assumption for learning theory, for practical usage this is difficult to ascertain. In deployed models, the incoming stream of data might start to be significantly different than the static dataset that the model was trained on -- a phenomenon named \textit{Concept drift} \cite{gama2014survey,wang2015concept}.  Formally, concept or data drift is defined as the scenario where the distribution of the data $P(X)$, the label $P(Y)$ or the concept $P(Y|X)$ changes as compared to the training data that the model has seen. Previous drift detection methods have either focused on the overall error rate \cite{gama2004learning}, or some other combination of the confusion matrix \cite{wang2013concept}, either of which require the prediction labels, which is not guaranteed for a machine learning model in production. Other work suggests using prediction drift as a proxy for concept drift in such cases \cite{vzliobaite2010change}. It informs of the change in the model's prediction distribution, and this information may be of importance to the practitioner, even if the model's accuracy is not impacted. For example, a lending company may have a quarterly target of loans to be disbursed, which may be achieved in a month, if the prediction distribution of real world applicants differs from training data, thereby causing issues in business planning. A systematic method is thus needed for studying prediction drift and attributing it to a) the features of the model and b) the individual data points that constitute the distributional samples that are compared. We frame the question as follows: \textit{Has the empirical distribution of inputs to the model drifted in a way that affects model behavior? If so, which features and which points in the sample have caused this shift?}

For this attribution to features and data, we focus on Shapley value based methods \cite{vstrumbelj2014explaining, lundberg2017unified,sundararajan2020many, datta2016algorithmic}. We include Integrated Gradients \cite{sundararajan2017axiomatic} in this broad family as it is equivalent to the Aumann-Shapley cost sharing method. Here, we adapt the Shapley framework, in the context of machine learning, for the following task: given two data samples of the same shape, and a function $D$ which computes some metric of distributional difference on the predictions made on the given datasets by a model $F$, attribute the output of $D$ to each point of the target dataset, and to each feature. By using the Shapley framework, we automatically inherit the Shapley axioms that have certain desirable properties, which we discuss in section \ref{sec:properties}.

Currently, there is no consensus on which of the many distributional difference metrics should be used to for calculating prediction drift, with previous work using measures like Jensen Shannon divergence \cite{pinto2019automatic}, Kolmogorov–Smirnov test \cite{dos2016fast} or the Wasserstein-1 distance \cite{miroshnikov2020wasserstein}. A comparative analysis of these methods is presented in Section \ref{sec:compmetrics}. We demonstrate an axiomatic framework to choose the most appropriate distributional distance metric depending on the use case.

\paragraph{Contributions}
Our key contributions in this paper are:
\begin{itemize}
    \item Establishing an axiomatic framework for calculating and explaining prediction drift using Shapley values and IG
    \item Extending the framework to explain arbitrary ``groups'' i.e. data and features together, thereby unifying several existing explanation methods
    \item Applying the Shapley values formulation to a function of distributional difference
    \item Axiomatization of measures of distributional difference
    \item Empirical analysis of the implications of choosing a particular metric of distributional difference to measure prediction drift, over a few handcrafted examples
\end{itemize}

\section{Related Work}

\paragraph{Concept Drift} The problem of concept drift in machine learning has been extensively studied in the literature -- spanning both sudden/instantaneous drift \cite{sudden} or slow/gradual drift \cite{stanley2003learning}. Furthermore, the literature makes a distinction between ``true" concept drift, and the ``virtual" concept drift that happens due to a change in data distribution, essentially a sampling issue \cite{salganicoff1997tolerating}.

In the literature, popular methods for detecting concept drift are AD-WIN \cite{bifet2007learning} and the Page-Hinkley Test \cite{gama2013evaluating}, with both these methods assuming that labels are available for analysis, which is infeasible in a scenario where the model is deployed and constantly making predictions on new data.

\paragraph{Prediction Drift as a proxy} In the absence of instantly available labels, other methods devolve to a measurement of the drift of the distributions of predictions as an ad-hoc method to detect concept drift. Methods using this approach include work by \citet{pinto2019automatic}, \citet{DBLP:journals/corr/abs-1902-02808}, \citet{dos2016fast} and \citet{vzliobaite2010change}. All these methods utilize different metrics to measure the difference in the distributions of the predictions of the new data points against a standard distribution.

% \paragraph{Distribution difference metrics} The prediction drift measuring methods above all utilize different metrics to measure the difference in the distributions of the predictions of the new data points against a standard distribution. Several 

\paragraph{Model explanations} Methods to describe the contribution of input features towards the final value of the prediction have gained considerable interest in the present, both from researchers and practitioners. One class of these methods effectively utilize Shapley value \cite{shapley1953value}, a popular concept in game theory to measure the contribution of each feature. Another method, Integrated Gradients \cite{sundararajan2017axiomatic}, utilizes the path between the input and a baseline for each feature to measure the attribution of each feature on that path, as a specialized case of the Shapley-like cost sharing method (commonly referred to as Shap). These methods have grown in popularity for quantifying the impact of features on a prediction at an instance level \cite{vstrumbelj2014explaining, datta2016algorithmic, lundberg2017unified}, on the loss at a global level \cite{covert2020understanding}, as also for quantifying the contribution of individual data points to a model's performance \cite{ghorbani2020distributional}. They have also been proposed to understand feature importance for measures of fairness \cite{begley2020explainability, miroshnikov2020wasserstein}.

\section{Terminology}

In this section, we lay out the terminology and notations that we use throughout the paper. 
% We re-use some terms from existing papers for easy understanding and to avoid confusion.
\begin{description}
    \item [Model function] – the machine learning model function $F$, which takes a vector of features of shape $n$ and returns an output(s). We limit the analysis to feature vectors instead of more general feature tensors, to avoid complications in notation. This does not mean however that the theory is particular to only models that accept single dimensional vectors, and can be extended quite easily.
    
    Similarly, for the sake of simplicity and without loss of generality, for models which output a vector of values, we analyze only one output at a time. For example, classification models output a vector of length equal to the number of classes, of which there is a particular class of interest which we wish to analyze.
    $F:R^{n} \to R$. Akin to machine learning models with batch predict, $F$ is also able to accept a batch input of shape $(m,n)$ and return m outputs.
    
    \item [Sample (of points)] – in the context of a model, a sample of $m$ feature vectors of shape $n$, the complete sample hence being of shape $(m, n)$. The sample could be a single point ($m=1$) or multiple ($m>1$). It could be chosen randomly from a distribution, or could be chosen intentionally as per requirements, e.g. points corresponding to men over the age of 50 from New York or the feature vector corresponding to ID ``x" in a database of customers of an online retail store.
    
    \item [Explicand] – the input sample of shape $(m,n)$ for which we want to explain the predictions, with respect to a particular model function.
    
    \item [Baseline] – a sample with the same dimensions as the explicand, against which the explicand is explained. All Shapley value based methods have a baseline, though it may not be obvious due to being implicit in the formulation \cite{sundararajan2020many,lundberg2017unified}. The explanation is dependent on the choice of baseline, and various papers \cite{merrick2020explanation} have proposed certain choices of baselines, or ways to select one. 
    
    \item [Value Function] – the set function $v(s)$, $v:2^{n} \to R$ that is used in the Shapley value formulation to obtain the attribution of each player. Here $n$ is the number of features, and $2^{n}$ refers to all possible combinations of feature presence (or absence).
    
    \item [Drift] – a measure of distributional difference, commonly used in context of time dependence, but we use it in a general sense.
    
    \item [Distributional drift function] – a function that given two samples (as defined above), returns a value characterizing the difference between them. We restrict ourselves to analyzing distributional differences over 1-D samples.
    
    $D(s_{1}, s_{2}): R^{n} \times R^{n} \to R$
    
    \item [Groups] – combinations of the $n \times m$ feature--data-point components belonging to the explicand.  These groups play the role of ``players'' in co-operative game theory for the purpose of Shapley and IG attributions of the drift value. Groups can be defined semantically, for example - males and females, and can be formed as combinations both in the feature and data-point dimensions. The Shapley value is calculated on the marginals of the resulting $m' \times n'$ groups as players that enter the coalition, over all such possible permutations. 
\end{description}

\section{Axioms}
\subsection{Axioms for attributions}

From  \cite{sundararajan2020many, friedman1999three}, we have the following desirable properties for attribution methods.  In Section \ref{sec:group-formulation} we will formulate GroupShapley and GroupIG such that they are inherited. We state them here in terms of the group formulation for convenience. Reasons for their desirability are expanded on in the Appendix.

\begin{enumerate}

    \item \textbf{Dummy} - this axiom states that a group that doesn't contribute to the game payout should get zero attribution.
    \item \textbf{Efficiency} - the sum of the attributions over all groups is equal to the difference of the model function's output at the explicand and the baseline. 
    \item \textbf{Linearity} - the attributions of the linear combination of the two model functions, are the same linear combination of the attributions of the model functions, taken one at a time. 
    \item \textbf{Symmetry} - for model functions that are symmetric for two groups $i$ and $j$, and the groups have the same value in both the explicand and baseline i.e. $g_{i} = g_{j}$ and $g'_{i} = g'_{j}$, the attributions to both the groups should be the same. 
    \item \textbf{Affine Scale Invariance} - requires the attributions to be invariant under the same affine transformation of both the model functions, and the groups.
    \item \textbf{Demand Monotonicity} - for a model function that is monotonic for a group, the attribution of the group should only increase if the value of the group increases.
    \item \textbf{Proportionality} - if the model function can be expressed as an additive sum of the input groups, and the baseline is zero, the attributions to each group are proportional to the group value.
\end{enumerate}

\subsection{Axioms for distributional drift functions}\label{sec:properties}

\citet{miroshnikov2020wasserstein} propose some desirable properties for a distributional drift function: 
\begin{enumerate}
  \item It should be \textbf{continuous} with respect to the change in the geometry of the distributions.
  \item It should be \textbf{non-invariant} with respect to monotone transformations of the distributions.
\end{enumerate}

Since our focus is on the distributional samples, and not the distributions themselves, we restate these properties for distributional drift measures $D$ for two 1-D samples. 

\begin{enumerate}
  \item \textbf{Sensitivity} - the drift function should be continuous with respect to changes in the individual points in the samples. For example, given two 1-D samples $S_1$ and $S_2$, if we change the value of any point in either, the function output should change.
  \item \textbf{Differentiability} - the drift function should be differentiable with respect to the individual points in the samples - this is a stronger version of the continuity axiom
  \item \textbf{Symmetry} - the drift function of two samples $S_1$ and $S_2$ should be symmetric i.e. $D(S_1, S_2) = D(S_2, S_1)$
  \item \textbf{Identity of Indiscernibles} - the drift is zero if and only if both samples are the same $D(S_1, S_1) = 0$ and $D(S_1, S_2) \neq 0$ if $S_1 \neq S_2$
  \item \textbf{Directionality} - the drift is signed based on the sample order $D(S_1, S_2) =  -D(S_2, S_1)$. A metric cannnot satisfy both Symmetry and Directionality, unless it's always zero
\end{enumerate}

\section{Prediction Drift}

We define prediction drift as the change in the distrbiution of the predictions of a model between two semantically meaningful slices of data.

The need for studying prediction drift to answer the question raised above 
arises due to the following reasons:

\begin{enumerate}

\item Detecting drift in the distribution of individual features may not be sufficient. For instance, it could be that the predictions may drift despite no drift in any of the individual feature distributions. This is because the joint distributions of the features may have drift.

\begin{table}[H]
    \centering
    \resizebox{\columnwidth}{!}{%
    \begin{tabular}{ c | c | c | c ||  c | c | c | c  }
    \toprule
    \multicolumn{4}{c||}{\bfseries Reference Distribution} & \multicolumn{4}{|c}{\bfseries Target Distribution}\\ \hline
	\textbf{x} & \textbf{y} & \textbf{z} & \textbf{f(x, y, z)} & \textbf{x} & \textbf{y} & \textbf{z} & \textbf{f(x, y, z)} \\ \midrule \midrule
    1.0 & 1.0 & 1.0 & 3.0 & 3.0 & 1.0 & 2.0 & 9.0 \\ 
	2.0 & 2.0 & 2.0 & 8.0 & 1.0 & 2.0 & 3.0 & 8.0 \\
	3.0 & 3.0 & 3.0 & 15.0 & 2.0 & 3.0 & 1.0 & 6.0 \\ \bottomrule
% 		1.0 & 1.0 & 1.0 & 3.0 & 3.0 & 6.0 & 3.0 & 18.0 \\ 
% 	2.0 & 2.0 & 2.0 & 8.0 & 9.0 & 7.0 & 1.0 & 17.0 \\
% 	3.0 & 3.0 & 3.0 & 15.0 & 7.0 & 5.0 & 5.0 & 45.0 \\ 
% 	4.0 & 4.0 & 4.0 & 24.0 & 10.0 & 8.0 & 7.0 & 85.0 \\
% 	5.0 & 5.0 & 5.0 & 35.0 & 2.0 & 4.0 & 9.0 & 31.0 \\ 
% 	6.0 & 6.0 & 6.0 & 48.0 & 1.0 & 10.0 & 6.0 & 22.0 \\ 
% 	7.0 & 7.0 & 7.0 & 63.0 & 8.0 & 2.0 & 8.0 & 74.0 \\
% 	8.0 & 8.0 & 8.0 & 80.0 & 6.0 & 9.0 & 4.0 & 37.0 \\ 
% 	9.0 & 9.0 & 9.0 & 99.0 & 5.0 & 3.0 & 2.0 & 15.0 \\ 
% 	10.0 & 10.0 & 10.0 & 120.0 & 4.0 & 1.0 & 10.0 & 51.0 \\ \bottomrule
\end{tabular}%
}
    \caption{The model function is $f(x, y, z) = xz + y + z$. The $x$, $y$ and $z$ distributions are unchanged at the univariate level, but the multivariate distribution has changed, so has the prediction distribution.}
    \label{tab:tab1}
\end{table}

\item Furthermore, drift in individual features may not always lead to drift in predictions. This could, for instance, happen if the drifting feature is unimportant to the model.

% \begin{table}[H]
%     \centering
%     \resizebox{\columnwidth}{!}{%
%     \begin{tabular}{ c | c | c | c ||  c | c | c | c  }
%     \toprule
%     \multicolumn{4}{c||}{\bfseries Reference Distribution} & \multicolumn{4}{|c}{\bfseries Target Distribution}\\ \hline
% 	\textbf{x} & \textbf{y} & \textbf{z} & \textbf{f(x, y, z)} & \textbf{x} & \textbf{y} & \textbf{z} & \textbf{f(x, y, z)} \\ \midrule \midrule
% 	1.0 & 1.0 & 1.0 & 2.001 & 1.0 & 11.0 & 1.0 & 2.011 \\ 
% 	2.0 & 2.0 & 2.0 & 6.002 & 2.0 & 12.0 & 2.0 & 6.012 \\ 
% 	3.0 & 3.0 & 3.0 & 12.003 & 3.0 & 13.0 & 3.0 & 12.013 \\ \bottomrule
% % 	1.0 & 1.0 & 1.0 & 2.001 & 1.0 & 11.0 & 1.0 & 2.011 \\ 
% % 	2.0 & 2.0 & 2.0 & 6.002 & 2.0 & 12.0 & 2.0 & 6.012 \\ 
% % 	3.0 & 3.0 & 3.0 & 12.003 & 3.0 & 13.0 & 3.0 & 12.013 \\ 
% % 	4.0 & 4.0 & 4.0 & 20.004 & 4.0 & 14.0 & 4.0 & 20.014 \\ 
% % 	5.0 & 5.0 & 5.0 & 30.005 & 5.0 & 15.0 & 5.0 & 30.015 \\ 
% % 	6.0 & 6.0 & 6.0 & 42.006 & 6.0 & 16.0 & 6.0 & 42.016 \\ 
% % 	7.0 & 7.0 & 7.0 & 56.007 & 7.0 & 17.0 & 7.0 & 56.017 \\ 
% % 	8.0 & 8.0 & 8.0 & 72.008 & 8.0 & 18.0 & 8.0 & 72.018 \\ 
% % 	9.0 & 9.0 & 9.0 & 90.009 & 9.0 & 19.0 & 9.0 & 90.019 \\ 
% % 	10.0 & 10.0 & 10.0 & 110.01 & 10.0 & 20.0 & 10.0 & 110.02 \\ \bottomrule
% \end{tabular}%
% }
%     \caption{ Consider the distributions of data given below, and the model function is $f(x, y, z) = xz + 0.001*y + z$. The $x$, and $z$ distributions are unchanged at the univariate level, but y has shifted significantly. But y is unimportant to the model, so the prediction distribution is stable.}
%     \label{tab:tab2}
% \end{table}

\item Finally, detecting drift in the prediction distributions may not be sufficient either. For instance, while the predictions distributions may remain the same, it could still be that the input feature distributions have changed in a meaningful way that affects how the model reasons. Such a drift is still worth noting.
For instance, the camera that feeds a face detection model could rotate over time, due to hinge failure. A robust model will be able to handle the distortion of the image for a while before it fails. The prediction distribution will not change initially, but the feature attributions over the pixels regions will change, which can serve as an early warning system.

\end{enumerate}

We focus our attention on problems 1 and 2, leaving 3 for future work. To answer the aforementioned question, we rely on the following steps:
\begin{itemize}
    \item Measure prediction drift for the model given two slices of data
    \item Attribute the drift to meaningful groups in the data.
\end{itemize}

Possible meaningful groups could be features of the model, n-tile buckets of predictions, or rule-based slices such as males vs females. We need to be careful to ensure that the the number of observations in each slice is proportionally similar for each sample, to avoid statistical anomalies seen in Simpson's paradox. \cite{simpson}

 Practically, for calculating the prediction drift given two data samples of unequal and/or large size, we suggest a bootstrapping approach. We sample from the two empirical distributions for a given number of repetitions and calculate the expected value of the prediction drift and the attributions and obtain statistical confidence bounds.
 
\section{Group Shapley and Group IG Formulation}\label{sec:group-formulation}

\subsection{The Shapley value}
Model function is $F:R^{n} \to R$

The Shapley value of a player $i$, playing an n-player coalitional game with a payout function $v$ is defined as

\begin{equation}
    \varphi _{i}(v)=\sum _{S\subseteq N\setminus \{i\}}{\frac {|S|!\;(n-|S|-1)!}{n!}}(v(S\cup \{i\})-v(S))
\end{equation}

\subsection{Baseline Shapley}
Baseline Shapley \cite{sundararajan2020many} or BShap, takes a function $f$, an explicand $x$ and a baseline $x'$. 

The value or payout function is  $v(S) = f(x_{S}; x'_{N\setminus S})$

Here, the absence of a feature is modeled using the corresponding baseline value. BShap is equivalent to the Shapley-Shubik cost sharing method and satisfies the following axioms: Dummy, Linearity, Affine Scale Invariance, Demand Monotonicity, and Symmetry.

\subsection{Integrated Gradients}

The Integrated Gradients formulation is 

\begin{equation}
IG_{i} (x, x', f):= (x_{i} - x'_{i})\int_{\alpha = 0}^1 \frac{\partial {F(x'+\alpha \times (x-x'))}}{\partial {x_i}}d\alpha
\end{equation}

Integrated gradients is equivalent to the Aumann-Shapley cost sharing method for continuous functions.

Integrated Gradients satisfies the following axioms: Dummy, Linearity, Affine Scale Invariance, Proportionality, and Symmetry

\subsection{Drift Group Shapley}

We define Drift Group Shapley, or GroupShapley, as being parametrized by the following choices:
\begin{enumerate}
    \item Choice of the explicand $S_{1}$ of shape $(m,n)$
    \item Choice of the baseline  $S_{2}$ of the same shape as the explicand
    \item A model function $F:R^{n} \to R$ 
    \item Additional functions, the chain of which we call $G$, which return two real valued outputs of equal shape for both the explicand and the sample
    \item Choice of a distributional difference function $D$, that takes two equal shaped outputs of the function $G \circ F$ and returns a real valued output
\end{enumerate}

The group formulation is:
\begin{equation}
    \sum_{n=1}^{G} \varphi_{i} = D(G(F(S_{1})), F(S_{2})))
\end{equation}

In GroupShapley, we explain the drift between the $G \circ F$ output of the explicand and the baseline.
 The number of players is equal to the number of groups times the number of features. The number of groups is the number of sub-divisions across rows. If the whole sample is one group, the features are the only players. If we have a row as it's own group, we end up with number of rows $\times$ number of features groups to which we attribute the payout. To be precise, we are attributing the drift score to each group in the explicand, where a group is a cross section consisting of at least one row and at most all rows, and at least one feature or at most $n$ features.

To simulate for missingness of a player, we replace the group of interest, with it's aligned counterpart from the reference dataset, similar to notion of the baseline in BShap or IG.

We now propose to frame every existing Shapley formulation as a prediction drift between some aspect of the model's behavior at the explicand and the baseline. We re-frame the two questions as:

\begin{enumerate}
    \item Has the empirical distribution of inputs to the model drifted in a way that affects model behavior? becomes \textit{Is there a difference in groups between the explicand and the baseline that affects some aspect of model behavior?}
    \item If so, which features and which points in the sample have caused this shift? becomes \textit{If so, which groups have caused it?}
\end{enumerate}

We list the following existing methods which we attempt to bring under a common umbrella:

\begin{enumerate}
    \item \cite{merrick2020explanation} unifies BShap/KernelSHAP/QII, noting that the KernelSHAP (CES) and QII (RBShap) can be derived by taking the expectation of BShap over particular distributions, namely the input distribution for KernelSHAP and the joint marginal for QII. The approach in \cite{vstrumbelj2014explaining} is equivalent to kernelSHAP \cite{sundararajan2020many} 

    Therefore, we can consider KernelSHAP and QII to be the following case of GroupShapley:
    Explicand is of shape $(1, n)$, broadcast to $(m, n)$ where m is the size of the background sample over which the expectation is calculated. The groups are the $n$ features and the drift function is the expected value difference.  

    \item  SAGE \cite{covert2020understanding} is a global explanation method, where the aim is to attribute the loss of the model to the features, by suggesting that a feature whose removal increases the loss is more important. The loss is computed over a data sample of shape $(m,n)$. They propose using the conditional distribution as in CES in theory, but in practice use the marginal, as in RBShap. This is equivalent to GroupShapley on $n$ groups, broadcasting the row dimension to $m \times j$ where $j$ is the size of the background baseline sample the applicable distribution. The drift function is the expected value difference.
    \item Distributional Shapley \cite{ghorbani2020distributional} aims to find the value of a data point, given a model and an evaluation metric. There is no inherent concept of a baseline here, though we could trivially add a set of random data as the baseline. We can design $G$ so as to make the $G \circ F$ of the baseline to be zero. The drift function is the expected value difference between the accuracy on the explicand and the artificially created zero value accuracy of the baseline. We note that it may be more instructive to introduce the notion of a baseline here, so as to ground the value of a datum in more definite terms. For example, is the data from source A more informative than source B.
    \item In \cite{miroshnikov2020wasserstein}, they propose using Shapley values to explain the Wasserstein-1 distance between two prediction samples, each belonging to a class of a protected attribute like Gender, Race and so on. This is  directly analogous to our scheme.

\end{enumerate}

\subsection{Drift Group Integrated Gradients}

We define Drift Group IG, or GroupIG, as being parametrized by the following choices:
\begin{enumerate}
    \item Choice of the explicand $S_{1}$ of shape $(m,n)$
    \item Choice of the baseline  $S_{2}$ of the same shape as the explicand
    \item A model function $F:R^{n} \to R$ that is end-to-end differentiable with respect to the inputs
    \item Additional functions, the chain of which we call $G$, which return two real valued outputs of equal shape for both the explicand and the sample. $G$ has to be differentiable in terms of the individual samples
    \item Choice of a distributional difference function $D$, that takes two equal shaped outputs of the function $G \circ F$ and returns a real valued output. Again, $D$ has to be differentiable in terms of the original input samples 
\end{enumerate}

In GroupIG, we go from the baseline sample to the explicand in a straight line path. 
We can thus say that IG is a particular case of Drift Group IG, where m = 1, G is the identity function and the distributional difference function is the expected value difference. If we are using the Wasserstein-1 distance for a single input, we re-frame the function as the absolute distance between the prediction at the input and the baseline prediction.

\section{Distributional Distance Metrics}\label{sec:compmetrics}
 We now discuss the properties of some of the widely used distance metrics for distances between two 1-D samples $R_{1}$, $R_{2}$ of data of length $n$.
 
\subsection{Wasserstein-1 Distance}

The Wasserstein-1 distance, also called the Earth Mover's distance or Mallows distance, is a well known metric from optimal transport theory, and widely used in statistics and machine learning. The mathematical properties which aid its suitability for our task are discussed below, building on prior work \cite{kolouri2018sliced, miroshnikov2020wasserstein, jiang2020wasserstein}

For the case of two 1-D samples, which is the case we are focusing on, the $W_{p}$ distance is the 
 $L_{p}$ norm of the sorted samples. The Wasserstein-1 distance is the special case where  $p$ is 1.

$W_{p} = (\frac{1}{n} \sum\limits_{n=1}^{n} |R_{1}(i) - R_{2}(i)|^{p} )^\frac{1}{p}$. Hence for p = 1, it reduces to the mean of the L1 norm. \cite{levina2001earth}

 The $\text{W}_1$ distance for empirical samples satisfies the following distributional axioms: Sensitivity, Differentiability, Symmetry, and the Identity of Indiscernibles. (Proofs in Appendix)

\subsection{Expected value difference}

Expected value difference, can be understood simply as the difference in the Expected Value of two distributions. Given two samples, it's the difference in the mean. This is a very intuitive concept, and is the simplest measurement of distributional difference, corresponding to a change in the first order moment.

$E(R_{1}, R_{2}) = \frac{1}{n} \sum\limits_{n=1}^{n} (R_{1}(i) - R_{2}(i))$. 

 The Expected value distance for empirical samples satisfies the following distributional axioms:
 Sensitivity, Differentiability, and Directionality but not the Identity of Indiscernibles. (Proofs in Appendix)

\subsection{Jensen Shannon Divergence}

The Jensen Shannon Divergence (JSD) given two probability distributions P and Q is defined as 

$JSD(P, Q) = \frac{1}{2}(D(P || M) + D(Q || M)$ 

where $M = \frac{1}{2}(P + Q)$ and $D$ is the Kullback-Liebler divergence. 

While it is difficult to analyze JSD's behavior given empirical samples, we can see that it does not satisfy Sensitivity and Directionality. (Proofs in Appendix)

\subsection{Kolmogorov-Smirnov Test Statistic for Two Samples}

This is actually a test to determine if two empirical probability distributions differ, and yields a distance that is used as measure of distributional difference. \cite{dos2016fast}

The KS statistic distance is defined as

$D(P, Q) = sup_{x} |P(x) - Q(x)|$ 

where $P(x)$ and $Q(x)$ are the empirical Cumulative Distribution Functions (CDF) of $R_{1}$ and $R_{2}$ and $sup$ is the supremum.

The KS Statistic satisfies only the Symmetry and the Identity of Indiscernibles axiom. \cite{miroshnikov2020wasserstein}

\begin{table*}[t]
    \centering
    \resizebox{\textwidth}{!}{
    \begin{tabular}{  c | c | c | c | c | c | c | c | c  }
    \toprule
\textbf{Function(x, y, z)} & \textbf{Explicand
[x, y, z]} & \textbf{Baseline
[x, y, z]} & \textbf{Exp. value Difference} & \textbf{$\text{W}_1$ Distance} & \textbf{Shapley} & \textbf{$\text{W}_1$ Shapley} & \textbf{IG} & \textbf{$\text{W}_1$ IG} \\ \midrule \midrule
	$xy$ & [1, 2, 3] & [0, 0, 0] & 2.0 & 2.0 & [1. 1. 0.] & [1. 1. 0.] & [1. 1. 0.] & [1. 1. 0.] \\
	$x - y$ & [1, 2, 3] & [0, 0, 0] & -1.0 & 1.0 & [ 1. -2. 0.] & [0. 1. 0.] & [ 1. -2. 0.] & [-1. 2. 0.] \\ 
	$x + y - z$ & [1, 2, 3] & [0, 0, 0] & 0.0 & 0.0 & [ 1. 2. -3.] & [0. 0. 0.] & [ 1. 2. -3.] & [0. 0. 0.] \\ 
	$xy - z^2$ & [1, 2, 3] & [0, 0, 0] & -7.0 & 7.0 & [ 1. 1. -9.] & [-0.33 -0.33 7.67] & [ 1., 1., -9., 0.] & [-1., -1., 9.] \\ 
	$min(x, y)$ & [1, 2, 3] & [0, 0, 0] & 1.0 & 1.0 & [0.5 0.5 0. ] & [0.5 0.5 0.] & [1 0 0. ] & [1 0 0. ] \\ 
	$abs(x- y)$ & [1, 2, 3] & [0, 0, 0] & 1.0 & 1.0 & [0. 1. 0.] & [0. 1. 0.] & [-1 2 0] & [-1 2 0] \\ \bottomrule
    \end{tabular}
    }
    \caption{BShap and IG Attributions for functions using Expected Value Difference and Wasserstein-1. Note the sparser attributions using the Wasserstein distance}
    \label{tab:my_label}
\end{table*}

\section{The concept of Alignment}

Given the need of a baseline in the Shapley value and IG formulations, it is natural to ask what is the right baseline, given that the attributions will differ with the choice of baseline. This is one of the most important questions in explainability.\cite{sundararajan2017axiomatic} recommends choosing a baseline where the model's prediction is neutral. \cite{merrick2020explanation} argues for contrastive explanations, with justification from norm theory \cite{kahneman1986norm}.

In GroupShapley and GroupIG when using the $\text{W}_1$ drift function, we take the counterpart in the other sample as baseline, when both samples are aligned by their sorted prediction values. The $\text{W}_1$ distance is based on the concept of optimal transport, and hence, the intuition extends naturally to the flow from the attributions, which make up the prediction from one distribution to the other. 

For other drift metrics, there may not be a natural reason to align in any particular way. But the alignment of the $\text{W}_1$ distance still can be justified as comparing the most similar points in the two samples, if the prediction of model is viewed as a task specific dimensionality reduction. Fliptest \cite{black2020fliptest} uses a similar thought process for assessing individual fairness by creating counterfactuals via optimal transport.

The alternative, where no choice needs to be made, is to take the expectation over all possible alignments. 

\section{Analysis}

We now look at some practical examples of how the choice of drift function impacts the explanations.

\subsection{Simple Experiments}

We analyze BShap and IG for a few functions in Table \ref{tab:my_label}, using both the expected value difference and the $\text{W}_1$ distance. These are functions of three variables x, y, and z, the baseline for all is [0, 0, 0], and the explicand is [1,2,3].
For the function $x-y$, we see that the $\text{W}_1$ attributions are different for different for both BShap and IG. It seems that the $\text{W}_1$ drift function gives sparser attributions for BShap, by compressing the attributions for the features that act in opposite direction to the eventual predicted value. For instance, for $x-y$, $x$ is 1 and $y$ is 2, so the prediction is -1, and the $\text{W}_1$ distance from the baseline prediction is 1. The method gives all the attribution to $y$, as it has the sign of the prediction. We can see this behavior for $x + y - z$ and $xy - z^{2}$ as well. This is reminiscent of how the L1 norm sparsifies coefficients in ridge regression, but we make no claims of there being any analogy between the two. 

There is no reason to always prefer the explanation of one over the other, both can be justified in their own way and are a matter of choice, similar to how choosing a baseline is a choice depending on the question one is looking to answer.

\subsection{Case Study}

We now present a simple case study, to demonstrate how this might work in practice, by constructing a synthetic dataset.  This allows us to inject known and controlled drifts in order to evaluate the effectiveness of various methods at finding them.

We create a dataset of the following features:

\begin{enumerate}[noitemsep]
    \item Location - \{`Springfield', `Centerville'\} - 70:30 
    \item Education - \{`GRAD',`POST\_GRAD'\} - 80:20
    \item Experience - years - (0, 50)  - normally distributed
    \item Engineer Type - \{`Software',`Hardware'\} - 85:15
    \item Relevant Experience - years - (0, 50)  - normally distributed
\end{enumerate}

and ensure that experience $\geq$ relevant experience.

The model predicts an individual's salary from the features above, using the following formula:

\begin{equation*}
\begin{split}
\text{Salary} = 50,000 + 20,000 \times \text{location} + 20,000\\ \times \text{education}+ 5,000 \times \text{relevant\_experience}\\ + 100 \times \text{experience} + 10,000 \times \text{engineer\_type}
\end{split}
\end{equation*}

\begin{figure}[H]
\centering
\includegraphics[width=0.45\textwidth]{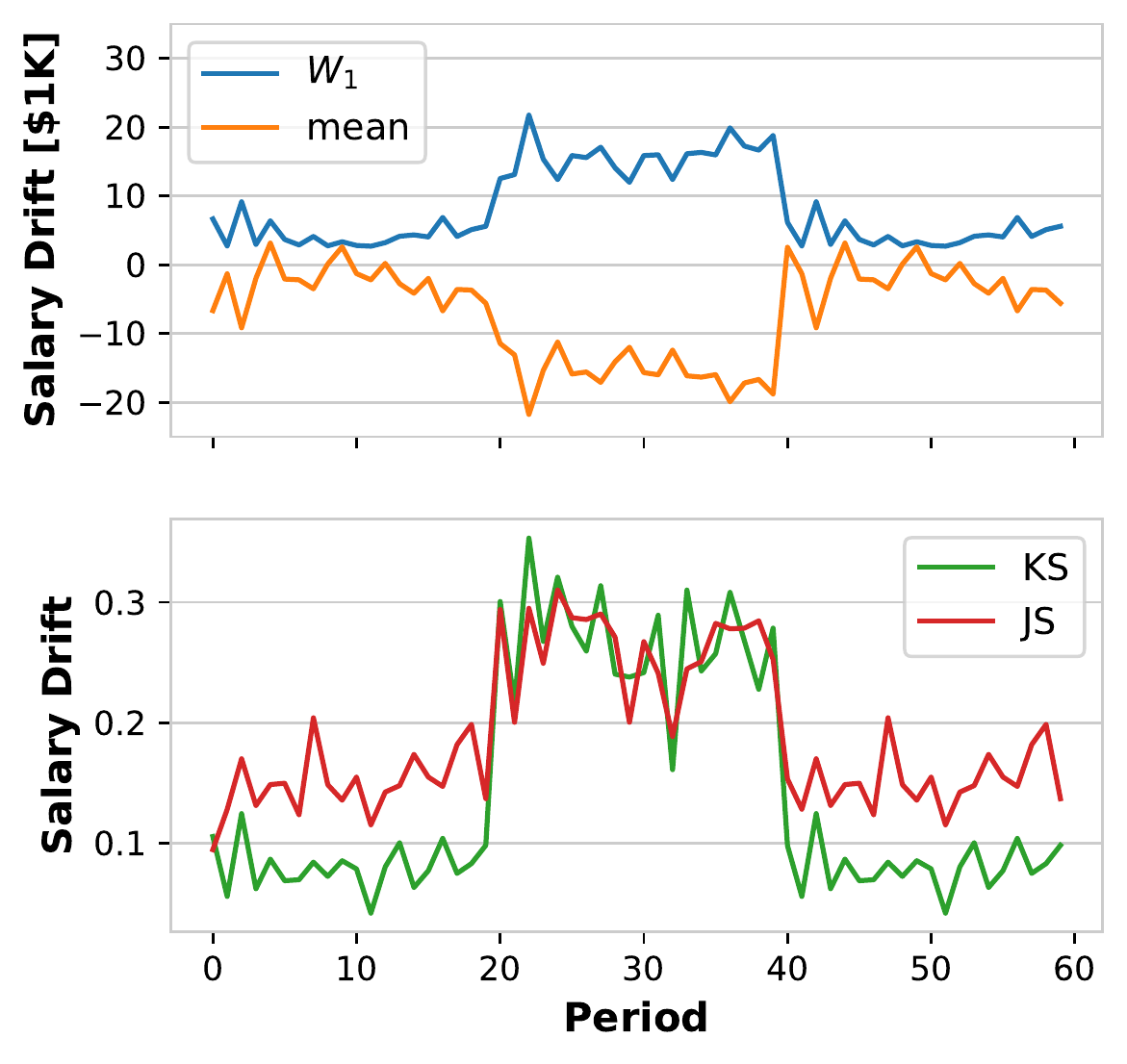}
\caption{Comparison of four distributional drift functions.  Wasserstein and means/expected value difference [upper] preserve the units of the predicted quantity and may provide a more intuitive scale.  The Kolmogorov-Smirnov Statistic and Jensen Shannon Divergence [lower] are dimensionless and the scale reflects the degree of absolute distributional overlap.  The central 20 periods have a data integrity error intentionally introduced which causes some applicants to have their location misinterpreted.}
\label{fig:Drift}
\end{figure}

2000 events are created for each of three days. On the second day, a plausible data pipeline bug is introduced, whereby the location feature has the value ``springfield'' rather than ``Springfield''. Because of this, all locations are identified as `Centerville', which leads to an average salary drop for day two–  a prediction drift. We now would like to attribute this the offending features. Figure \ref{fig:Drift} shows the drift over time measured by the various drift methods previously discussed.

\begin{figure}[t]
\centering
\includegraphics[width=0.45\textwidth]{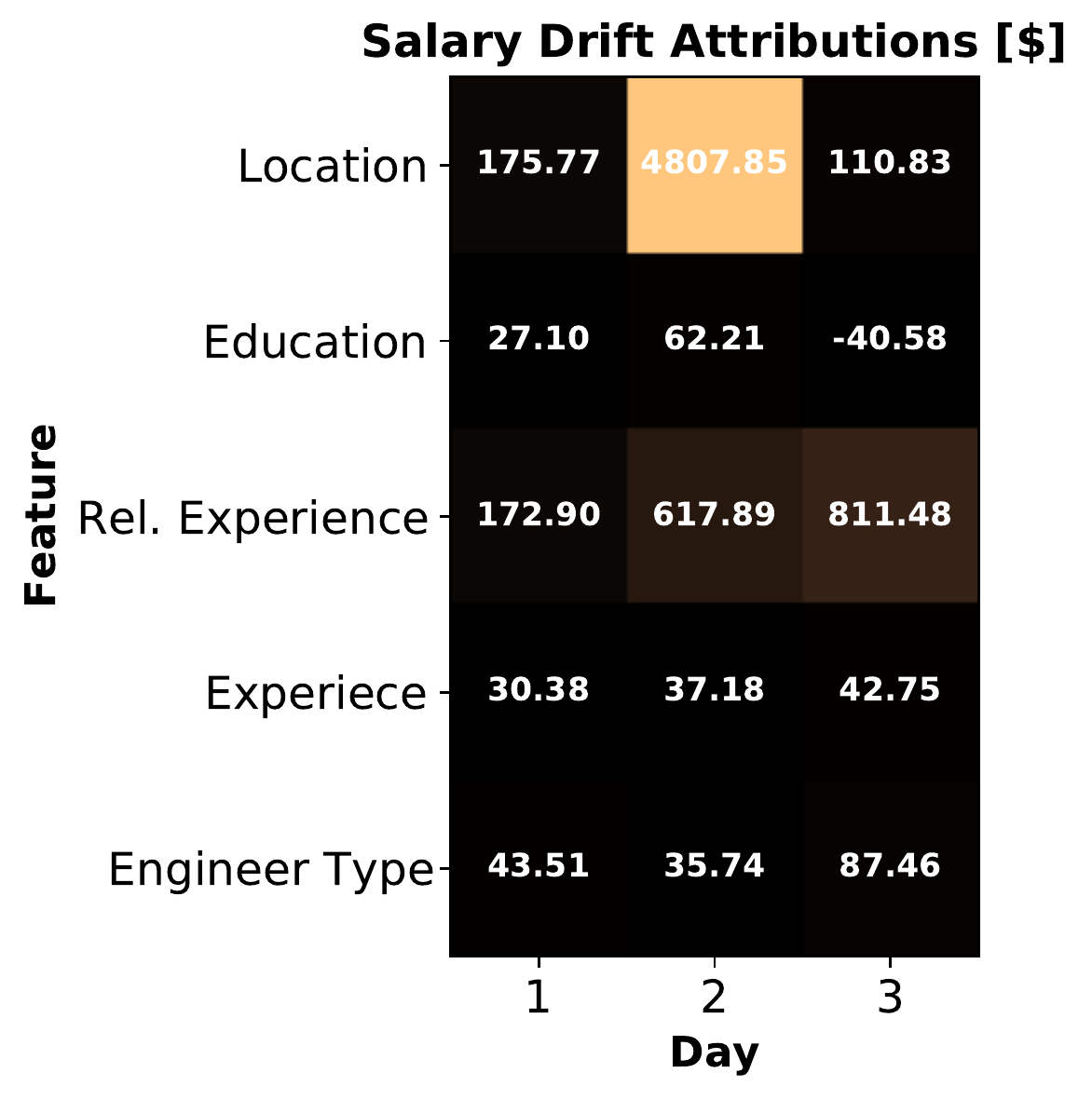}
\caption{$W_1$ measures a drift of \$7022.46 over the complete dataset.  By forming groups of [features]$\times$[days], GroupShapley unambiguously identifies the source of drift as the ``Location'' feature on the second day.}
\label{fig:W1}
\end{figure}

In Figure \ref{fig:W1}, we calculate GroupShapley attributions over the fifteen feature-day combinations, and see that the job location feature gets the most attribution, as we would expect.

\begin{figure}[t]
\centering
\includegraphics[width=0.48\textwidth]{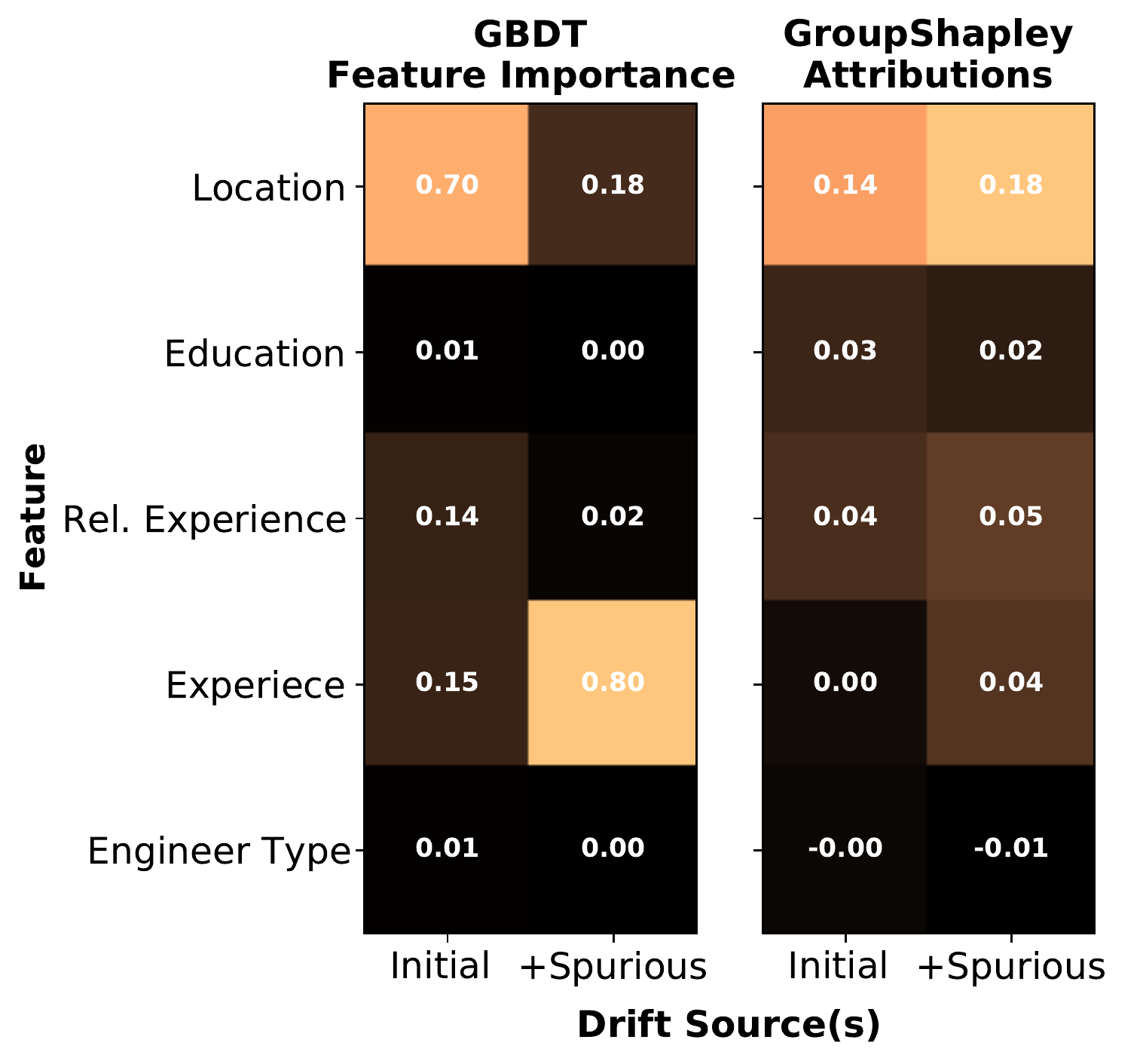}
\caption{Comparison of tree-based feature importance method from \cite{pinto2019automatic} and GroupShapley.  Both methods initially identify the correct source of drift; but when an additional correlated feature drift is added, the GBDT method assigns it most of the importance, despite its minimal effect on the model output.}
\label{fig:GBDT}
\end{figure}

Additionally, we compare our approach to that of \cite{pinto2019automatic}, which measures drift using Jensen Shannon divergence and trains a Gradient Boosted Tree Classifier to identify the drift.  The feature importances of the classifier are used to identify the cause.  In the scenario described, it correctly gives the most attribution to the location feature. But if we introduce another spurious drift, of an unimportant feature like experience, the GBDT method selects the wrong feature. They do suggest a technique to remove time-trended features, but if the other feature also spikes in the same interval, that fix will not help either as seen in Figure \ref{fig:GBDT}.

% \begin{figure}[H]
% \centering
% \includegraphics[width=0.5\textwidth]{rugplot.pdf}
% \caption{Caption}
% \end{figure}

% \begin{figure}[H]
% \centering
% \includegraphics[width=0.5\textwidth]{wass.pdf}
% \caption{Caption}
% \end{figure}

% \begin{figure}[H]
% \centering
% \includegraphics[width=0.5\textwidth]{jsd.pdf}
% \caption{Caption}
% \end{figure}

% \begin{figure}[H]
% \centering
% \includegraphics[width=0.5\textwidth]{ks.pdf}
% \caption{Caption}
% \end{figure}

\section{Conclusion and Future Work}
We study the problem of prediction drift and attributing it, and propose it as a general framework of explainability, unifying several methods. We axiomatize certain desirable properties of distributional difference metrics, also demonstrating that explanation methods can be parameterized by the choice of this metric.

 A more detailed study of the theoretical implications of choosing one distance metric over another for explanations is left for future work. Additionally, GroupShapley can be computationally expensive, and approximation schemes for faster calculations could be a future area of exploration.

\clearpage

\section{Appendix}
\subsection{Axioms}

We will now go over the reasons for the desirability of the axioms:

\begin{enumerate}

    \item \textbf{Dummy} - We do not want to credit a group/feature that makes no contribution to the model prediction.
    \item \textbf{Efficiency} - This ensures a complete accounting of difference in the model's prediction between the explicand and the baseline. 
    \item \textbf{Linearity} - This property helps in avoiding counter-intuitive behavior when analyzing attributions of linear functions.
    \item \textbf{Symmetry} - The purpose of this axiom is self-evident, if two groups contribute equally they should receive the same attribution.
    \item \textbf{Affine Scale Invariance} - The justification for this is based on the idea that the units of measurement of individual features may not be comparable to each other, and secondly, within themselves, may not be canonical. For example, units of weight like pounds or kilograms are not more or less justified than the other, and the conversion to the other should not lead to a decrease in attribution.  \cite{friedman1999three}
    \item \textbf{Demand Monotonicity} - For a function that is monotonic with respect to a group, if the group value increases while all else is held constant, the function's value will increase. It is natural to want the attribution to the group to increase as compared to the previous scenario. 
    \item \textbf{Proportionality} - This ensures that the attributions to groups are proportional to their contribution in the additive sum of the group values. Let's look at a heat generation scenario. If there are three current sources, each supplying the same amount of current. The heat generated is proportional to the square of the current. The attribution to each should be one-third, compared to the zero baseline. Now if we combine two of the current sources, the attribution of the third should ideally remain the same.
\end{enumerate}
 
\subsection{Proofs for Drift Metrics satisfying Axioms}
\textbf{Wasserstein-1 Distance}

Given two samples $R_{1}$ and $R_{2}$ of length $n$ and sorted by value, the $W_{1}$ distance can be computed $W_{1} = \frac{1}{n} \sum\limits_{i=1}^{n} \left|R_{1}(i) - R_{2}(i)\right|$. 

The $W_1$ distance for empirical samples satisfies the following distributional axioms:
 
\textbf{Proofs:} 

\begin{enumerate}
   \item \textbf{Sensitivity} - This is trivial to see, given that each point of the sample contributes to the overall sum.
   \item \textbf{Differentiability} - The function is piece-wise differentiable, except at zero for each absolute difference.
   \item \textbf{Symmetry} - The formula is symmetric in $R_{1}$ and $R_{2}$.
   \item \textbf{Identity of Indiscernibles} - The $W_1$ distance can be zero only if every element-pair in the two samples cancels each other out.
\end{enumerate}

\textbf{Expected value difference}

Given two samples $R_{1}$ and $R_{2}$ of length $n$, the Expected value distance is
$E(R_{1}, R_{2}) \approx \frac{1}{n} \sum\limits_{i=1}^{n} \left(R_{1}(i) - R_{2}(i)\right)$. 

\textbf{Proofs:} 

\begin{enumerate}
   \item \textbf{Sensitivity} - Each point of the sample contributes to the overall sum.
   \item \textbf{Differentiability} - One can see that the function is differentiable everywhere.
   \item \textbf{Directionality} - The sign changes when the sample order is flipped.
   \item \textbf{Identity of Indiscernibles} - This can be proved by a counter example. If there is a sample that only has values 1, and the other has equal number of zeros and twos. The two means will be equal and will cancel out, even though the two samples are not the same.
\end{enumerate}

\textbf{Jensen Shannon Divergence}

The Jensen Shannon Divergence (JSD) given two probability distributions P and Q is defined as $JSD(P, Q) = \frac{1}{2}(D(P || M) + D(Q || M)$ where $M = \frac{1}{2}(P + Q)$ and $D$ is the Kullback-Liebler divergence. 

While it is difficult to analyze JSD's behavior given empirical samples, we can see that it does not satisfy Sensitivity and Directionality.

\textbf{Proofs:} 

\begin{enumerate}
   \item \textbf{Sensitivity} - This can be proved by a counter example. If there are two distributions that don't intersect anywhere, the JSD is one. Now if we translate the second distribution while ensuring there is no intersection, the JSD is still 1.
   \item \textbf{Directionality} - JSD is symmetric to the change in the sample order.

\end{enumerate}
\textbf{Kolmogorov-Smirnov Test Statistic for Two Samples}

For two distributions $R_1$ and $R_2$, the KS statistic distance is $D(P, Q) = \sup_{x} |P(x) - Q(x)|$ where $P$ and $Q$ are the empirical Cumulative Distribution Functions (CDF) of $R_{1}$ and $R_{2}$

We can see from the definition that the KS Statistic satisfies Symmetry and the Identity of Indiscernibles axiom. For the other proofs please refer to \cite{miroshnikov2020wasserstein}
\clearpage

% % Acknowledgements should only appear in the accepted version.
% \section*{Acknowledgements}

% \textbf{Do not} include acknowledgements in the initial version of
% the paper submitted for blind review.

% If a paper is accepted, the final camera-ready version can (and
% probably should) include acknowledgements. In this case, please
% place such acknowledgements in an unnumbered section at the
% end of the paper. Typically, this will include thanks to reviewers
% who gave useful comments, to colleagues who contributed to the ideas,
% and to funding agencies and corporate sponsors that provided financial
% support.

% In the unusual situation where you want a paper to appear in the
% references without citing it in the main text, use \nocite

\bibliographystyle{icml2021}
\bibliography{ref.bib}

\begin{thebibliography}{31}
\providecommand{\natexlab}[1]{#1}
\providecommand{\url}[1]{\texttt{#1}}
\expandafter\ifx\csname urlstyle\endcsname\relax
  \providecommand{\doi}[1]{doi: #1}\else
  \providecommand{\doi}{doi: \begingroup \urlstyle{rm}\Url}\fi

\bibitem[Begley et~al.(2020)Begley, Schwedes, Frye, and
  Feige]{begley2020explainability}
Begley, T., Schwedes, T., Frye, C., and Feige, I.
\newblock Explainability for fair machine learning.
\newblock \emph{arXiv preprint arXiv:2010.07389}, 2020.

\bibitem[Bifet \& Gavalda(2007)Bifet and Gavalda]{bifet2007learning}
Bifet, A. and Gavalda, R.
\newblock Learning from time-changing data with adaptive windowing.
\newblock In \emph{Proceedings of the 2007 SIAM international conference on
  data mining}, pp.\  443--448. SIAM, 2007.

\bibitem[Black et~al.(2020)Black, Yeom, and Fredrikson]{black2020fliptest}
Black, E., Yeom, S., and Fredrikson, M.
\newblock Fliptest: fairness testing via optimal transport.
\newblock In \emph{Proceedings of the 2020 Conference on Fairness,
  Accountability, and Transparency}, pp.\  111--121, 2020.

\bibitem[Covert et~al.(2020)Covert, Lundberg, and Lee]{covert2020understanding}
Covert, I., Lundberg, S., and Lee, S.-I.
\newblock Understanding global feature contributions with additive importance
  measures.
\newblock \emph{Advances in Neural Information Processing Systems}, 33, 2020.

\bibitem[Datta et~al.(2016)Datta, Sen, and Zick]{datta2016algorithmic}
Datta, A., Sen, S., and Zick, Y.
\newblock Algorithmic transparency via quantitative input influence: Theory and
  experiments with learning systems.
\newblock In \emph{2016 IEEE symposium on security and privacy (SP)}, pp.\
  598--617. IEEE, 2016.

\bibitem[dos Reis et~al.(2016)dos Reis, Flach, Matwin, and
  Batista]{dos2016fast}
dos Reis, D.~M., Flach, P., Matwin, S., and Batista, G.
\newblock Fast unsupervised online drift detection using incremental
  kolmogorov-smirnov test.
\newblock In \emph{Proceedings of the 22nd ACM SIGKDD International Conference
  on Knowledge Discovery and Data Mining}, pp.\  1545--1554, 2016.

\bibitem[Friedman \& Moulin(1999)Friedman and Moulin]{friedman1999three}
Friedman, E. and Moulin, H.
\newblock Three methods to share joint costs or surplus.
\newblock \emph{Journal of economic Theory}, 87\penalty0 (2):\penalty0
  275--312, 1999.

\bibitem[Gama et~al.(2004)Gama, Medas, Castillo, and
  Rodrigues]{gama2004learning}
Gama, J., Medas, P., Castillo, G., and Rodrigues, P.
\newblock Learning with drift detection.
\newblock In \emph{Brazilian symposium on artificial intelligence}, pp.\
  286--295. Springer, 2004.

\bibitem[Gama et~al.(2013)Gama, Sebastiao, and Rodrigues]{gama2013evaluating}
Gama, J., Sebastiao, R., and Rodrigues, P.~P.
\newblock On evaluating stream learning algorithms.
\newblock \emph{Machine learning}, 90\penalty0 (3):\penalty0 317--346, 2013.

\bibitem[Gama et~al.(2014)Gama, {\v{Z}}liobait{\.e}, Bifet, Pechenizkiy, and
  Bouchachia]{gama2014survey}
Gama, J., {\v{Z}}liobait{\.e}, I., Bifet, A., Pechenizkiy, M., and Bouchachia,
  A.
\newblock A survey on concept drift adaptation.
\newblock \emph{ACM computing surveys (CSUR)}, 46\penalty0 (4):\penalty0 1--37,
  2014.

\bibitem[Ghanta et~al.(2019)Ghanta, Subramanian, Khermosh, Sundararaman, Shah,
  Goldberg, Roselli, and Talagala]{DBLP:journals/corr/abs-1902-02808}
Ghanta, S., Subramanian, S., Khermosh, L., Sundararaman, S., Shah, H.,
  Goldberg, Y., Roselli, D.~S., and Talagala, N.
\newblock {ML} health: Fitness tracking for production models.
\newblock \emph{CoRR}, abs/1902.02808, 2019.
\newblock URL \url{http://arxiv.org/abs/1902.02808}.

\bibitem[Ghorbani et~al.(2020)Ghorbani, Kim, and
  Zou]{ghorbani2020distributional}
Ghorbani, A., Kim, M., and Zou, J.
\newblock A distributional framework for data valuation.
\newblock In \emph{International Conference on Machine Learning}, pp.\
  3535--3544. PMLR, 2020.

\bibitem[Jiang et~al.(2020)Jiang, Pacchiano, Stepleton, Jiang, and
  Chiappa]{jiang2020wasserstein}
Jiang, R., Pacchiano, A., Stepleton, T., Jiang, H., and Chiappa, S.
\newblock Wasserstein fair classification.
\newblock In \emph{Uncertainty in Artificial Intelligence}, pp.\  862--872.
  PMLR, 2020.

\bibitem[Kahneman \& Miller(1986)Kahneman and Miller]{kahneman1986norm}
Kahneman, D. and Miller, D.~T.
\newblock Norm theory: Comparing reality to its alternatives.
\newblock \emph{Psychological review}, 93\penalty0 (2):\penalty0 136, 1986.

\bibitem[Kolouri et~al.(2018)Kolouri, Pope, Martin, and
  Rohde]{kolouri2018sliced}
Kolouri, S., Pope, P.~E., Martin, C.~E., and Rohde, G.~K.
\newblock Sliced-wasserstein autoencoder: An embarrassingly simple generative
  model.
\newblock \emph{arXiv preprint arXiv:1804.01947}, 2018.

\bibitem[Levina \& Bickel(2001)Levina and Bickel]{levina2001earth}
Levina, E. and Bickel, P.
\newblock The earth mover's distance is the mallows distance: Some insights
  from statistics.
\newblock In \emph{Proceedings Eighth IEEE International Conference on Computer
  Vision. ICCV 2001}, volume~2, pp.\  251--256. IEEE, 2001.

\bibitem[Lundberg \& Lee(2017)Lundberg and Lee]{lundberg2017unified}
Lundberg, S. and Lee, S.-I.
\newblock A unified approach to interpreting model predictions.
\newblock \emph{arXiv preprint arXiv:1705.07874}, 2017.

\bibitem[Merrick \& Taly(2020)Merrick and Taly]{merrick2020explanation}
Merrick, L. and Taly, A.
\newblock The explanation game: Explaining machine learning models using
  shapley values.
\newblock In \emph{International Cross-Domain Conference for Machine Learning
  and Knowledge Extraction}, pp.\  17--38. Springer, 2020.

\bibitem[Miroshnikov et~al.(2020)Miroshnikov, Kotsiopoulos, Franks, and
  Kannan]{miroshnikov2020wasserstein}
Miroshnikov, A., Kotsiopoulos, K., Franks, R., and Kannan, A.~R.
\newblock Wasserstein-based fairness interpretability framework for machine
  learning models.
\newblock \emph{arXiv preprint arXiv:2011.03156}, 2020.

\bibitem[{Nishida} et~al.(2008){Nishida}, {Shimada}, {Ishikawa}, and
  {Yamauchi}]{sudden}
{Nishida}, K., {Shimada}, S., {Ishikawa}, S., and {Yamauchi}, K.
\newblock Detecting sudden concept drift with knowledge of human behavior.
\newblock In \emph{2008 IEEE International Conference on Systems, Man and
  Cybernetics}, pp.\  3261--3267, 2008.
\newblock \doi{10.1109/ICSMC.2008.4811799}.

\bibitem[Pinto et~al.(2019)Pinto, Sampaio, and Bizarro]{pinto2019automatic}
Pinto, F., Sampaio, M.~O., and Bizarro, P.
\newblock Automatic model monitoring for data streams.
\newblock \emph{arXiv preprint arXiv:1908.04240}, 2019.

\bibitem[Salganicoff(1997)]{salganicoff1997tolerating}
Salganicoff, M.
\newblock Tolerating concept and sampling shift in lazy learning using
  prediction error context switching.
\newblock In \emph{Lazy learning}, pp.\  133--155. Springer, 1997.

\bibitem[Shapley(1953)]{shapley1953value}
Shapley, L.~S.
\newblock A value for n-person games, contributions to the theory of games ii
  (aw tucker and hw kuhn, eds.), 1953.

\bibitem[Stanley(2003)]{stanley2003learning}
Stanley, K.~O.
\newblock Learning concept drift with a committee of decision trees.
\newblock \emph{Informe t{\'e}cnico: UT-AI-TR-03-302, Department of Computer
  Sciences, University of Texas at Austin, USA}, 2003.

\bibitem[{\v{S}}trumbelj \& Kononenko(2014){\v{S}}trumbelj and
  Kononenko]{vstrumbelj2014explaining}
{\v{S}}trumbelj, E. and Kononenko, I.
\newblock Explaining prediction models and individual predictions with feature
  contributions.
\newblock \emph{Knowledge and information systems}, 41\penalty0 (3):\penalty0
  647--665, 2014.

\bibitem[Sundararajan \& Najmi(2020)Sundararajan and
  Najmi]{sundararajan2020many}
Sundararajan, M. and Najmi, A.
\newblock The many shapley values for model explanation.
\newblock In \emph{International Conference on Machine Learning}, pp.\
  9269--9278. PMLR, 2020.

\bibitem[Sundararajan et~al.(2017)Sundararajan, Taly, and
  Yan]{sundararajan2017axiomatic}
Sundararajan, M., Taly, A., and Yan, Q.
\newblock Axiomatic attribution for deep networks.
\newblock In \emph{International Conference on Machine Learning}, pp.\
  3319--3328. PMLR, 2017.

\bibitem[Wagner(1982)]{simpson}
Wagner, C.~H.
\newblock Simpson's paradox in real life.
\newblock \emph{The American Statistician}, 36\penalty0 (1):\penalty0 46--48,
  1982.
\newblock ISSN 00031305.
\newblock URL \url{http://www.jstor.org/stable/2684093}.

\bibitem[Wang \& Abraham(2015)Wang and Abraham]{wang2015concept}
Wang, H. and Abraham, Z.
\newblock Concept drift detection for streaming data.
\newblock In \emph{2015 international joint conference on neural networks
  (IJCNN)}, pp.\  1--9. IEEE, 2015.

\bibitem[Wang et~al.(2013)Wang, Minku, Ghezzi, Caltabiano, Tino, and
  Yao]{wang2013concept}
Wang, S., Minku, L.~L., Ghezzi, D., Caltabiano, D., Tino, P., and Yao, X.
\newblock Concept drift detection for online class imbalance learning.
\newblock In \emph{The 2013 International Joint Conference on Neural Networks
  (IJCNN)}, pp.\  1--10. IEEE, 2013.

\bibitem[{\v{Z}}liobaite(2010)]{vzliobaite2010change}
{\v{Z}}liobaite, I.
\newblock Change with delayed labeling: When is it detectable?
\newblock In \emph{2010 IEEE International Conference on Data Mining
  Workshops}, pp.\  843--850. IEEE, 2010.

\end{thebibliography}

%%%%%%%%%%%%%%%%%%%%%%%%%%%%%%%%%%%%%%%%%%%%%%%%%%%%%%%%%%%%%%%%%%%%%%%%%%%%%%%
%%%%%%%%%%%%%%%%%%%%%%%%%%%%%%%%%%%%%%%%%%%%%%%%%%%%%%%%%%%%%%%%%%%%%%%%%%%%%%%
% DELETE THIS PART. DO NOT PLACE CONTENT AFTER THE REFERENCES!
%%%%%%%%%%%%%%%%%%%%%%%%%%%%%%%%%%%%%%%%%%%%%%%%%%%%%%%%%%%%%%%%%%%%%%%%%%%%%%%
%%%%%%%%%%%%%%%%%%%%%%%%%%%%%%%%%%%%%%%%%%%%%%%%%%%%%%%%%%%%%%%%%%%%%%%%%%%%%%%

% \appendix
% \section{Do \emph{not} have an appendix here}

% \textbf{\emph{Do not put content after the references.}}
% %
% Put anything that you might normally include after the references in a separate
% supplementary file.

% We recommend that you build supplementary material in a separate document.
% If you must create one PDF and cut it up, please be careful to use a tool that
% doesn't alter the margins, and that doesn't aggressively rewrite the PDF file.
% pdftk usually works fine. 

% \textbf{Please do not use Apple's preview to cut off supplementary material.} In
% previous years it has altered margins, and created headaches at the camera-ready
% stage. 

%%%%%%%%%%%%%%%%%%%%%%%%%%%%%%%%%%%%%%%%%%%%%%%%%%%%%%%%%%%%%%%%%%%%%%%%%%%%%%%
%%%%%%%%%%%%%%%%%%%%%%%%%%%%%%%%%%%%%%%%%%%%%%%%%%%%%%%%%%%%%%%%%%%%%%%%%%%%%%%

\end{document}